%% file: main.tex
\documentclass[letterpaper]{article} 
\usepackage[]{paper}  
\nocopyright
\usepackage{times}  
\usepackage{helvet}  
\usepackage{courier}  
\usepackage[hyphens]{url}  
\usepackage{graphicx} 
\usepackage{natbib}  
\usepackage{caption} 
\usepackage{algorithm}
\usepackage{algorithmic}
\usepackage{newfloat}
\usepackage{listings}
\DeclareCaptionStyle{ruled}{labelfont=normalfont,labelsep=colon,strut=off} 
\floatstyle{ruled}
\newfloat{listing}{tb}{lst}{}
\floatname{listing}{Listing}
\usepackage{multirow}
\usepackage{booktabs}
\usepackage{amsthm}

\usepackage{amsfonts}
\usepackage{amsmath}
\usepackage[textsize=tiny]{todonotes}
\definecolor{coolgrey}{rgb}{0.55, 0.57, 0.67}

\newcommand{\zhouztd}[1]

\title{ Enabling Small Models for Zero-Shot Selection and Reuse\\through Model Label Learning}
\author{
Jia Zhang\textsuperscript{\rm 12\dag}, 
Zhi Zhou\textsuperscript{\rm 1\dag}, 
Lan-Zhe Guo\textsuperscript{\rm 13\ddag}, 
Yu-Feng Li\textsuperscript{\rm 12\ddag}}
\affiliations{
    \textsuperscript{\rm 1}National Key Laboratory for Novel Software Technology, Nanjing University
    
    \textsuperscript{\rm 2}School of Artifical Intelligence, Nanjing University

    \textsuperscript{\rm 3}School of Intelligence Science and Technology, Nanjing University

    \texttt{\{zhangjia, zhouz\}@lamda.nju.edu.cn}

    \textsuperscript{\dag} Equal Contribution \textsuperscript{\ddag} Corresponding Author
    
}

\begin{document}

\maketitle

\renewcommand\thefootnote{}
\footnotetext{Preprint. Under review.}

\begin{abstract}

Vision-language models (VLMs) like CLIP have demonstrated impressive zero-shot ability in image classification tasks by aligning text and images but suffer inferior performance compared with task-specific expert models. 
On the contrary, expert models excel in their specialized domains but lack zero-shot ability for new tasks. How to obtain both the high performance of expert models and zero-shot ability is an important research direction. 
In this paper, we attempt to demonstrate that by constructing a model hub and aligning models with their functionalities using model labels, new tasks can be solved in a zero-shot manner by effectively selecting and reusing models in the hub. 
We introduce a novel paradigm, Model Label Learning (MLL), which bridges the gap between models and their functionalities through a Semantic Directed Acyclic Graph (SDAG) and leverages an algorithm, Classification Head Combination Optimization (CHCO), to select capable models for new tasks. 
Compared with the foundation model paradigm, it is less costly and more scalable, i.e., the zero-shot ability grows with the sizes of the model hub. Experiments on seven real-world datasets validate the effectiveness and efficiency of MLL, demonstrating that expert models can be effectively reused for zero-shot tasks. Our code will be released publicly.

\end{abstract}

\section{Introduction}
\input{sections/introduction}

\section{Related Work}
\input{sections/related_work}

\section{Problem and Analysis}
\input{sections/problem_formulation}

\section{Methodology}
\input{sections/methodology}

\section{Experiments}
\input{sections/experiments}

\section{Conclusion}

In this paper, we explore the area of reusing task-specific expert models for zero-shot classification tasks. We present the Model Label Learning (MLL) paradigm to align models in the model hub with their functionalities using model labels, enabling effective model reuse for zero-shot tasks. We make a preliminary attempt to implement the MLL
paradigm, consisting of three key steps: model labelling, model selection, and model reuse. This paradigm is less costly and more scalable than VLMs. Experiments validate the feasibility and effectiveness of MLL, demonstrating that expert models can be effectively selected and reused for zero-shot prediction.

One limitation is that the current SDAG used for labeling model is a discrete structure. Developing more flexible model labelling methods leaves for future research.

\appendix

\label{sec:reference}
\bibliography{ref}

\end{document}

%% file: sections/introduction.tex




Zero-shot classification~\cite{wang2019survey} has gained significant attention due to its ability to bypass the need for extra model training and user-provided task-specific training data, leading to more user-friendly machine learning applications. 
Recently emerged vision-language models (VLMs), such as CLIP~\cite{radford2021learning}, have demonstrated impressive zero-shot abilities in image classification by training on large-scale image-text paired datasets to align images and text in a joint embedding space.

\input{sections/figures/figure_intro1}


Though VLMs show promising zero-shot abilities, it is reported that VLMs suffer inferior performance~\cite{mo2023sclip, saha2024improved} compared with task-specific expert models on downstream tasks. Furthermore, training large VLMs consumes a significant amount of computing and data resources, and their scalability is constrained due to the difficulty of updating such extensive models. On the contrary, we can easily obtain various expert models for different tasks in the machine learning community; for example, more than eight hundred thousand models are available in HuggingFace. These models excel in their respective specialized tasks but are deficient in zero-shot learning for novel tasks. Therefore, it is natural to consider
whether we can
organize these task-specific expert models to enable them with zero-shot capability.



To this end, we propose a novel paradigm called Model Label Learning (MLL).
Our basic idea is to assign each model a label that describes its specific functionality, including the classes it excels at classifying and to what extent.
Leveraging model labels enables the selection and reuse of models to make predictions on test data in a zero-shot manner for downstream tasks. 
Compared with VLMs, whose zero-shot capacity lies in the alignment between text and images, aligning models with their functionalities using model labels enables the effective selection and reuse of models for handling new tasks.
Figure~\ref{fig:pic1} illustrates the comparison between VLMs and MLL.
Moreover, the MLL paradigm is less costly and more scalable since the zero-shot classification capability grows with the size of the model hub. 


In this paper, we make a preliminary attempt to implement the MLL paradigm with three steps: model labelling, model selection, and model reuse. 
Specifically, for model labelling, we construct a Semantic Directed Acyclic Graph (SDAG) where each node describes a semantic class with corresponding representative data. 
We pre-test each model to generate a model label that describes how well it excels at predicting these semantic classes. 
For model selection, model labels first semantically match a preliminary candidate model set for target classes when a downstream task comes.
Then, we propose a novel method called Classification Heads Combination Optimization (CHCO) to select capable models for target tasks further. 
Finally, for model reuse, we assemble selected models using computed ensemble weights
to handle the target task in a zero-shot manner. 
Extensive experimental results demonstrate that expert models can be effectively selected and reused for zero-shot classification with the proposal, and the scalability of the model hub has also been proved.

To summarize, our contributions are as follows:
\begin{enumerate}
    \item[(1)] We introduce a novel paradigm, model label learning, to enable task-specific expert models for zero-shot classification tasks by aligning models with their functionalities using model labels.    This paradigm is helpful in building less costly and more scalable methods with zero-shot classification capability compared to VLMs.
    \item[(2)] We make a preliminary attempt to implement the MLL paradigm in three key steps: model labelling, model selection, and model reuse. This method demonstrates the feasibility of MLL and could facilitate related research.
    \item[(3)] Experiments validate the effectiveness and efficiency of MLL, demonstrating that expert models can be effectively reused in a zero-shot manner.

\end{enumerate}


%% file: sections/figures/figure_intro1.tex
\begin{figure}[t]
    \centering
    \includegraphics[width=\columnwidth]{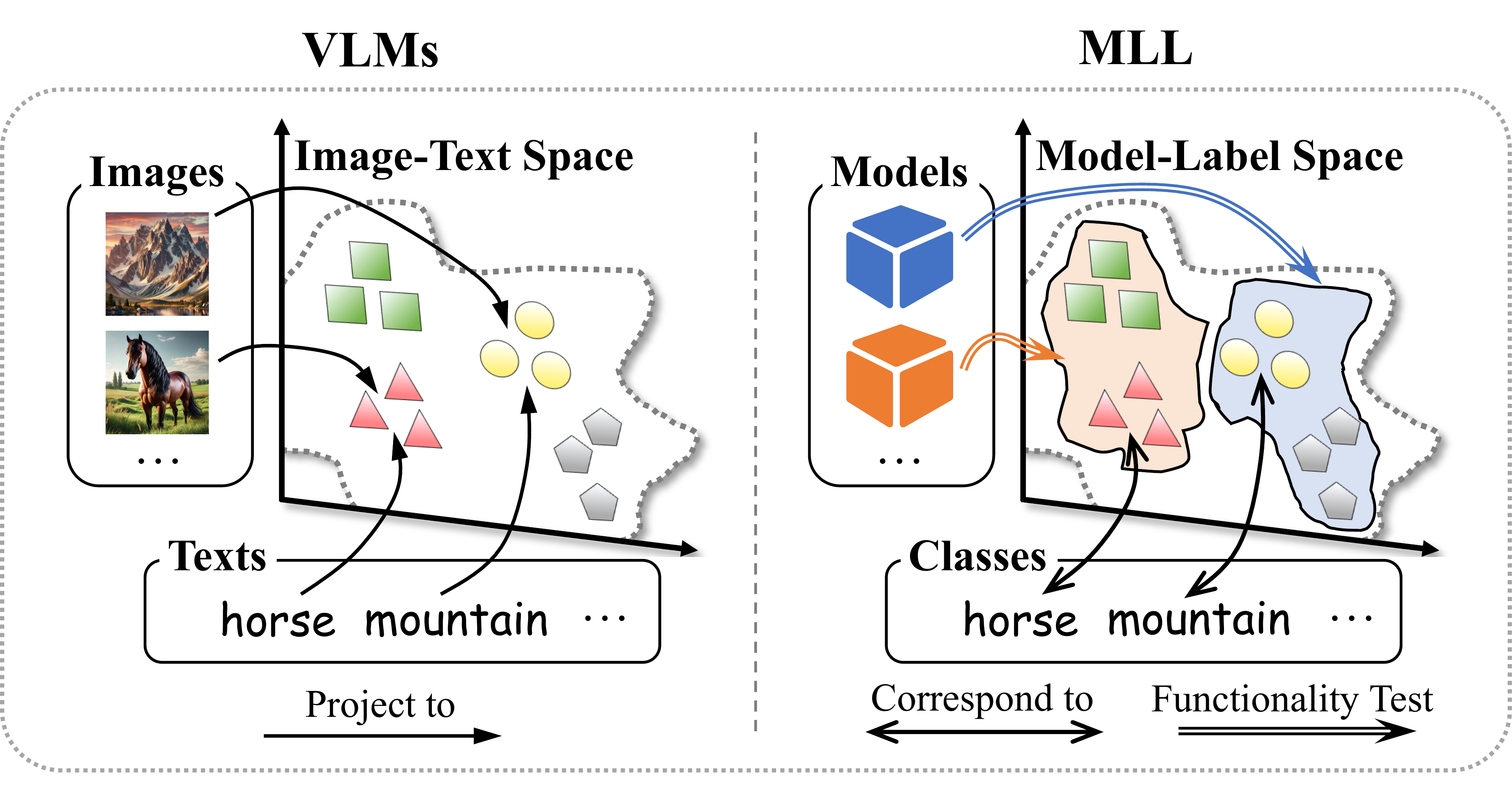} 
    \caption{The comparison between VLMs and MLL. MLL aligns models in the hub with their functionalities using model labels to identify capable expert models for tasks.}
    \label{fig:pic1}
\end{figure}

%% file: sections/related_work.tex
The goal of the proposal is to enable task-specific expert models with zero-shot ability, and the main technique is to organize a model hub to enable model selection and reuse. Thus, we review zero-shot learning and model selection methods related to our work.

\subsection{Zero-Shot Learning}
Zero-shot~\cite{wang2019survey} is a machine learning paradigm requiring the model to recognize unseen classes without additional training data. This is typically achieved by leveraging semantic information, such as attributes or textual descriptions, bridging between known and unknown classes. Traditional techniques~\cite{palatucci2009zeroshot, lampert2013attributebased, frome2013devise} rely on pre-defined attributes or textual data to map the relationship between seen and unseen classes. Recently, VLMs have demonstrated impressive zero-shot capabilities by training on vast text-image paired data to align texts and images in a common embedding space. For instance, CLIP~\cite{radford2021learning} employs a vision and a language model to learn joint embeddings of images and text through contrastive learning with extensive paired data, ALIGN~\cite{jia2021scaling} trains with a contrastive loss on a large dataset of noisy image-text pairs, FLAVA~\cite{singh2022flava} learns from both paired and unpaired images and texts using various losses. Given a new image, the model classifies the image by matching it with a text description such as ``a photo of [class]" and selecting the most related classes in the image-text embedding space.

\subsection{Model Selection}
In an era rich with various pre-trained models and model hubs, many practitioners are trying to exploit the enormous pre-trained models to help solve AI tasks ~\cite{han2021pretrained, renggli2022which}. However, how to select the appropriate model for a new task relied on experience and brute-force fine-tuning of various candidate models to identify the best one. Various reusability assessment methods have been proposed to enhance the efficiency of model selection, such as NCE ~\cite{tran2019transferability}, LEEP ~\cite{nguyen2020leep}, and LogME ~\cite{you2021logme}. These methods assess the compatibility between the target task data and features generated by candidate models and then select models that are more likely to achieve better reuse performance on the target task. Although more efficient than brute-force fine-tuning, these approaches still require forward inference on the target task for all candidate models. Emerging paradigms, such as learnware ~\cite{zhou2016learnware, guo2023identifying, tan2024enabling}, explore more effective model selection. In this paradigm, numerous models are organized within a learnware docker, each accompanied by a training data distribution-driven specification describing its functionality~\cite{wu2021model}. Users can search for suitable models based on specifications and their data distribution. Although these works have attempted to build model hubs to organize models, they have not enabled the model hub to have zero-shot classification capabilities.

%% file: sections/problem_formulation.tex
This section first outlines the notions and problem formulation for model label learning and then analyzes the three key steps in the proposal.

\input{sections/figures/figure_method1}

\subsection{Problem Formulation}

We focus on solving zero-shot multi-class classification tasks using expert models. We are provided a model hub \(\mathcal{M}=\{F_m\}_{m=1}^{M}\) containing $M$ expert models, each trained well on a specific domain. We denote $H_m$ as the number of output dimension for each model $F_m$, such that $F_m(x) \in \mathbb{R}^{H_m}$. As $H_m$ varies across models, their output spaces are heterogeneous. Additionally, each expert model is uploaded anonymously, and we have no knowledge regarding its training data distribution, architectures, functionalities, or output space semantics. 
The target zero-shot classification task involves an input space $\mathcal{X}$ and candidate task classes $\mathcal{Y}=\{y\}$, where each $y$ represents the textual description of a class. In the zero-shot setting, no training data is available. 
We aim to select and reuse some useful expert models from the hub to achieve zero-shot classification for the test data.

Enabling expert models for zero-shot tasks via MLL has three key steps: model labelling, model selection, and model reuse. Labelling involves assigning functionality labels to models, while selection entails assessing task-specific reuse scores based on model labels and identifying capable experts. The model reuse step requires the ensemble of selected models to make predictions in a zero-shot manner. 

\subsection{Analysis of Model Labelling}
Just as a hirer's selection of candidates relies on tests and appraisals rather than just resumes, model labelling relies on pre-testing for better insight into the model's functionality.
A semantic-based data structure should be constructed to pre-test expert models and assign them labels that describe how well it excels at classifying some semantic classes.
Although we attempt to produce labels describing functionality, it is inaccurate to measure the model's ability to recognize certain classes using fixed reuse scores without knowing the target task in advance. For instance, a model well-trained on a dog-vs-cat dataset may be able to distinguish dogs from leopards, but it might not perform well on the dog-vs-wolf task. Task-specific reuse scores depend on both the model and the target task. Thus, the labelling method should preprocess enough functionality descriptions as model labels to facilitate the matching of candidate models and target classes and the assessment of the task-specific reuse score once the target task is given.

\subsection{Analysis of Model Selection and Reuse}
To achieve model reuse for the zero-shot prediction, the target task on $\mathcal{Y}$ is treated as $|\mathcal{Y}|$ one-vs-rest classification tasks. We have to select one-vs-rest expert predictors and generate task-specific reuse scores to identify useful ones. 
Although a model with \(H_m\) classification heads can be intuitively viewed as \(H_m\) one-vs-rest predictors, constructing each binary predictor using only one head does not work that well. In practical reuse scenarios, the heterogeneous relationship between the target class and the model output head is typically complex rather than simple one-to-one mappings. Combining the model's multiple classification heads to form a predictor for collective recognition yields better performance. However, searching through various possible combinations of heads poses a challenge.

%% file: sections/figures/figure_method1.tex
\begin{figure*}[t]
    \centering
    \includegraphics[width=0.9\textwidth]{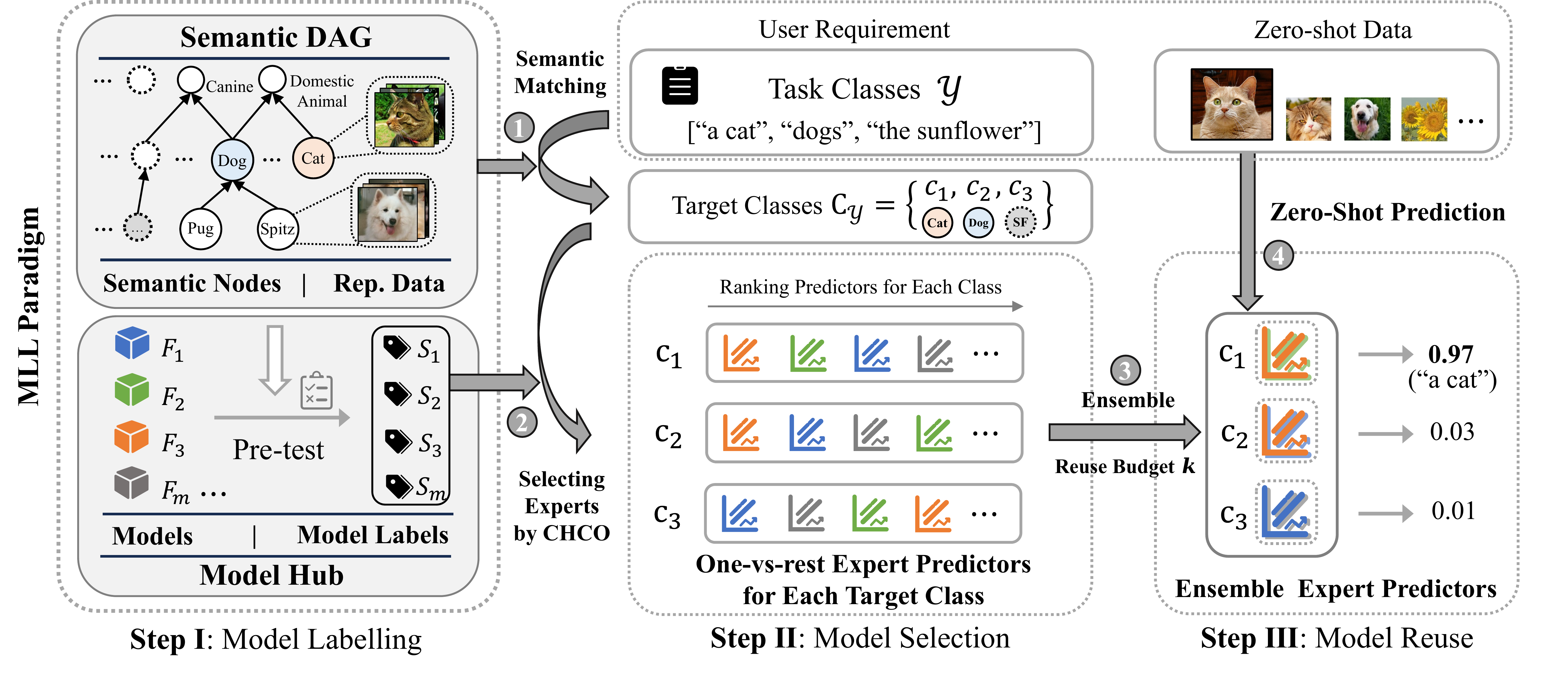} 
    \caption{The overview framework of the MLL paradigm. Models submitted to the hub undergo pre-testing to receive model labels that describe their functionalities in the labelling step. When a user's downstream task arises, the proposal selects useful experts in the selection step and assembles them to handle the task in a zero-shot manner.}
    \label{fig:pic2}
\end{figure*}

%% file: sections/methodology.tex
We make a preliminary attempt to implement the MLL paradigm with three key steps: model labelling, model selection, and model reuse. We first outline the overall framework and then delve into the details of each step.

\subsection{Overall Framework}
As illustrated in Figure \ref{fig:pic2}, the MLL system comprises a Semantic Directed Acyclic Graph (SDAG) and a model hub. During the labelling step, models submitted to the hub are pre-tested and assigned labels $S_m$ describing their functionalities using the SDAG, where each node describes a semantic class with corresponding representative data. In the selection step, the text classes of the target task \(\mathcal{Y}\) are mapped into the target classes \(C_\mathcal{Y}\) by textual semantic matching with the SDAG. We then propose the Classification Heads Combination Optimization algorithm (CHCO), utilizing model labels $S_m$ to generate task-specific reuse scores and select one-vs-rest expert predictors. Given a specified reuse budget \(k\), we assemble up to \(k\) predictors for each class. Larger reuse budgets intuitively lead to better predictions but increase costs.
These \(|\mathcal{Y}|\) ensemble expert predictors \(\{F_c^k\}\) are employed in the reuse step to handle the target task in a zero-shot manner.

\subsection{Step \uppercase\expandafter{\romannumeral1}. Model Labelling}
We build a preliminary Semantic Directed Acyclic Graph, (SDAG), containing semantic class nodes \( C = \{c\} \) and the corresponding representative data \( D_c = \{x\} \). By leveraging the semantic alignment between images in ImageNet~\cite{deng2009imagenet}  and WordNet ~\cite{miller1995wordnet} textual synsets, each node \(c\) is associated with a WordNet noun synset, with a few images randomly sampled from ImageNet as \( D_c\) to represent the actual image data distribution of this semantics. However, a few sampled images are insufficient to capture the full distribution for nodes with broader semantics such as \textit{Domestic Dog} or \textit{Animal}. Thus, we further structure the nodes into a Directed Acyclic Graph (DAG) structure based on the hypernym-hyponym relationships of synsets in WordNet. By employing topological ordering and a discount factor $\delta$, we leverage hierarchical semantic relationships to integrate more comprehensive testing results, thereby reducing the sampling cost and enhancing the utilization of representative data in the SDAG.

\begin{equation}
    \mathrm{logit}^c_m=\frac{1}{|D_c|}\sum_{x\in{D_c}}F_m^\mathrm{logit}(x),
\end{equation}
\begin{equation}
    s^c_m=\delta \cdot \mathrm{logit}^{c}_m + \frac{(1-\delta)}{|\Omega(c)|}\sum_{c'\in \Omega(c)} \mathrm{logit}^{c'}_m,
\end{equation}
\begin{equation}
    S_m = \{s^c_m|c\in C\},
\end{equation}
where \(F_m^\mathrm{logit}(x)\) represents the logits vector output by model \( F_m \) given \( x \) and $\Omega(c)$ denotes the set of first-order successors of node $c$. Finally, We assign \(S_m\) to $F_m$ as the model label.

The MLL paradigm supports other model labelling methods to capture functionalities. For instance, reusability assessment methods such as LogME or LEEP can be employed to generate metrics as model labels, describing how well the models excel at recognizing semantic classes. When the downstream task comes, these labels serve as reuse scores directly to select models. Detailed descriptions and comparisons are provided in experiments.
Additionally, the SDAG is scalable. New nodes with clear relationships to existing nodes can be inserted into the SDAG; otherwise, they can be added as independent nodes. After adding new nodes or modifying the representative data of existing nodes, models can be quickly tested on the incremental data, and the model labels updated accordingly.

\input{sections/tables/table1}

\subsection{Step \uppercase\expandafter{\romannumeral2}. Model Selection}
For model selection, we first map all text classes of target task \(\mathcal{Y}\) to target classes \(C_\mathcal{Y}\) through semantic matching with nodes in the SDAG. The matching process is performed using the cosine similarity of text embeddings generated by a pre-trained language model. We then propose the Classification Head Combination Optimization (CHCO) algorithm to select useful expert predictors for each class in \(C_\mathcal{Y}\).

\vspace{5pt}
\noindent \textbf{CHCO Algorithm}. To simplify the notation, we omit the subscript \( m \) in the following discussion. Consider a model has \( |H| \) classification heads, denoted as \( H = \{ h \} \). For the target classes \( C_\mathcal{Y} = \{ c \} \), we aim to construct \( |\mathcal{Y}| \) binary predictors from the model's classification heads to optimize the expected performance over \( C_\mathcal{Y} \). We focus on the subset \( S_\mathcal{Y} = \{ s^c \mid c \in C_\mathcal{Y} \} \) of the model label \( S \), where \( s^c = \{ s_h^c \} \) represents the average logit output by classification head \( h \) for class \( c \). The head-class probability matrix \( P = (p_{hc}) \in \mathbb{R}^{|H| \times |\mathcal{Y}|} \) is compute based on \( s^c \):

\begin{align}
    p_{hc}=\frac{\exp(s^c_h)}{\sum_{h'}\exp(s_{h'}^c)}
    \label{eq:st0}
\end{align}

A binary predictor \( F_c \) for class \(c\) can be represented as a linear combination of multiple heads. We define the head combination variable \( X = (x_{hc}) \in \mathbb{R}^{|H| \times |\mathcal{Y}|} \), where \( x_{hc} \) represents the proportion of head \( h \) used for predicting class \( c \). \( X \) should satisfy the following constraints:
\begin{equation}
    \sum_{c\in C_\mathcal{Y}} x_{hc} \leq 1,\ \forall\ h\in H
    \label{eq:st1}
\end{equation}
\begin{equation}
    0 \leq x_{hc} \leq 1,\ \forall\ h\in H,c\in C_\mathcal{Y}
    \label{eq:st2}
\end{equation}

\input{sections/algorithms/algorithm2}

Based on \( X \) and \( P \), we could further compute the expected probability output of each binary predictor for class \( c \) and other classes \( c'\in C_\mathcal{Y}/\{c\}\) as \(p_c(c)\) and \(p_c(c')\), respectively. We then derive the negative log-likelihood discriminative loss \(\mathcal{L}_c\) for each predictor \( F_c \):

\begin{equation}
    p_c(c)=\sum_{h\in H} p_{hc}\cdot x_{hc},\ \forall c \in C_\mathcal{Y}
    \label{eq:st3}
\end{equation}
\begin{equation}
    p_c(c')=\sum_{h\in H} p_{hc'}\cdot x_{hc},\ \forall c \in C_\mathcal{Y}, c'\in C_\mathcal{Y}/\{c\}
    \label{eq:st4}
\end{equation}
\begin{small}
\begin{equation}
\mathcal{L}_c=-\left[\log(p_c(c))+\frac{1}{|\mathcal{Y}|-1}\sum_{c' \in C_\mathcal{Y}/\{c\}}\log(1-p_c(c'))\right]
\label{eq:st5}
\end{equation}
\end{small}

Our objective is to minimize the loss across all classes while adhering to constraints on the head combinations:
\begin{align}
\underset{X}{\arg\min}\  \mathcal{L}=\underset{X}{\arg\min}\ \sum_c\mathcal{L}_c\\ 
\text{s.t.}\quad (\ref{eq:st0}),(\ref{eq:st1}),(\ref{eq:st2}),(\ref{eq:st3}),(\ref{eq:st4}),(\ref{eq:st5}) \notag
\end{align}

We achieve this optimization by alternately applying gradient descent and simplex projection. In each iteration, we select a head \( h \) and fix the combination variables \( x_{h'c} \) for the remaining heads \(h'\). We then take the derivative of the loss \( \mathcal{L} \) with respect to \( x_{hc} \) 
, yielding:

\begin{small}
    \begin{align}
    \frac{\partial \mathcal{L}}{x_{hc}} = \frac{\partial \mathcal{L}_c}{x_{hc}}
    &=\frac{1}{|\mathcal{Y}|-1}\sum_{c'\in C_\mathcal{Y}/\{c\}}\frac{p_{hc'}}{1-\sum_{h'\in H}p_{h'c'}x_{h'c}}\notag\\
    &-\frac{p_{hc}}{\sum_{h'\in H}p_{h'c}x_{h'c}},
\end{align}
\end{small}

\noindent which indicates that all combination variables \( x_{hc}\) of head \(h\) independently affect the corresponding loss \( \mathcal{L}_c \). Since \( \mathcal{L}_c \) is a convex function with respect to \( x_{hc} \) due to the convexity of the logarithmic function, optimizing \( \mathbf{x}_h \) along the gradients with an appropriate learning rate will ensure a decrease and convergence of \( \mathcal{L} \). To maintain constraints, after each gradient descent step, we project \( \mathbf{x}_h \) onto the simplex ~\cite{duchi2008efficient}. Iterating this  process until convergence yields the combination scheme for the model's classification heads.

For each model \( F_m \), we apply CHCO to solve for the combination variable \( X_m \) and calculate the loss \( \mathcal{L}_c \) of each of the \( |\mathcal{Y}| \) binary predictors as their task-specific reuse scores. We ultimately obtain \( M\) binary predictors for each class \( c \). Given a specified reuse budget \( k \), for each class \( c \), we select up to \( k \) top-scoring expert predictors with reuse scores below a threshold \(\beta\) to form an ensemble predictor \( F_c^k=\{(F_m,\ \mathbf{x}^m_c)\} \), where \(\mathbf{x}^m_c\) is the \(c\)-th column vector of \(X_m\). 

\subsection{Step \uppercase\expandafter{\romannumeral3}. Model Reuse}
Consequently, the model selection step yields \(|\mathcal{Y}|\) ensemble expert predictors for recognizing the classes within \(C_\mathcal{Y}\).  During testing, for any zero-shot data \( x \in \mathcal{X}\) of the user task, all ensemble expert predictors \(\{F_c^k\}\) are employed in a zero-shot manner to infer the confidence \( p_c^k(x) \), and the class with the highest confidence is chosen as the prediction \(y'\) for \( x \):

\begin{align}
\label{eq.predict}
    p^k_c(x) = \sum_{(F_m,\ \mathbf{x}^m_c) \in F_c^k} \omega_m^c \cdot F_m(x)^T \mathbf{x}^m_c,
\end{align}
\begin{align}
\label{eq.weights}
    \omega_m^c=\frac{1-\mathcal{H}(F_m(x))/\log{H_m}}{\sum_{F_{m'} \in F_c^k}(1-\mathcal{H}(F_{m'}(x))/\log{H_{m'}})},
\end{align}
\noindent where \(\mathcal{H}\) denotes the probability entropy and \(\omega_m^c\) denotes the ensemble weights obtained by normalizing output probability entropy of each expert model within \(F_c^k\) to reduce the influence of unreliable predictions. 
We summarize the model selection and reuse procedure in Algorithm \ref{alg:alg2} and the CHCO algorithm in the appendix.

\subsection{Interim MLL Patch towards Practice}
The MLL paradigm is built upon a fully developed MLL system, where the SDAG covers the task classes, and models in the hub manage them. However, prior to reaching this advanced stage, we must contend with an underdeveloped MLL system. Therefore, We implement an interim patch using a general VLM. For task classes that the MLL system cannot match or resolve, zero-shot predictions are handled by the general VLM. As the MLL system evolves, specialized experts will eventually replace the general big model. We present the results for both full and partial coverage of the target task by the MLL system in experiments.

\noindent\textbf{Proposition 4.1.} \textit{Given the decomposition of target task \(\mathcal{Y}=\mathcal{Y}^e\cup\mathcal{Y}^g\), where \(\mathcal{Y}^e\) and \(\mathcal{Y}^g\) represent for the classes handled by the reuse of experts and the general VLM, respectively. The expected errors of the general model, {\small\(E^g\)}, reusing experts through MLL, {\small\(E\)} and reusing experts after enhancing MLL system, {\small\(\tilde{E}\)}, satisfy that {\small\(\tilde{E} \leq E \leq E^g\)}.}

\noindent\textit{Remark 4.1.} Proposition 4.1 and the later experimental results demonstrates that, even when the target task calls for the assistance of a general VLM, the effective reuse of experts improves the performance compared to relying solely on a single general VLM. Additionally, evolving and scaling the MLL system lead to continuous improvement. The proof and more details refer to the appendix.

%% file: sections/tables/table1.tex
\begin{table*}[ht!]
\small
\centering
\caption{Comparison of the zero-shot performance on seven tasks. The table above reports the accuracy for the portion of tasks covered by the MLL system. The table below reports the accuracy of entire tasks. The best performance is highlighted in bold.}
\label{tab:exp1}
\begin{tabular}{@{}l|ccccccc|c@{}}
\toprule
\midrule
Acc.(c) & MA-DOG & MA-AWA & MA-BRD & Oxford-Pets & Food-101 & Caltech-101 & Flowers-102 & \quad Avg.\quad\quad \\ \midrule
CLIP & 83.42 & 94.08 & 92.00 & 87.40 & 82.36 & 89.04 & 67.20 & 85.07 \\
Random\textsuperscript{MLL} & 46.32 & 42.43 & 50.75 & 47.77 & 28.95 & 31.71 & 15.24 & 37.60 \\
LogME\textsuperscript{MLL} & 80.79 & 40.79 & 46.00 & 64.47 & 49.22 & 73.17 & 78.72 & 62.78 \\
LEEP\textsuperscript{MLL} & 95.92 & 86.97 & 71.25 & 92.05 & 41.54 & 80.02 & 43.16 & 72.99 \\
MLL & \textbf{96.97} & \textbf{97.50} & \textbf{100.0} & \textbf{98.61} & \textbf{95.63} & \textbf{94.35} & \textbf{97.55} & \textbf{97.23} \\ \midrule
Cvr. & 1.00 & 0.70 & 0.30 & 0.53 & 0.32 & 0.64 & 0.46 & 0.51 \\ \midrule \midrule
Acc.(a) & MA-DOG & MA-AWA & MA-BRD & Oxford-Pets & Food-101 & Caltech-101 & Flowers-102 & Avg. \\ \midrule
CLIP & 83.42 & 94.25 & 77.63 & 83.11 & 82.04 & 88.63 & 63.67 & 81.82 \\
Random\textsuperscript{MLL} & 46.32 & 41.37 & 51.68 & 56.43 & 59.99 & 43.20 & 32.37 & 47.34 \\
LogME\textsuperscript{MLL} & 82.90 & 37.63 & 44.63 & 68.23 & 67.69 & 75.70 & 59.47 & 64.53 \\
LEEP\textsuperscript{MLL} & 95.92 & 85.88 & 67.00 & 83.80 & 64.27 & 80.64 & 46.19 & 74.81 \\
MLL & \textbf{96.97} & \textbf{95.63} & \textbf{78.13} & \textbf{86.14} & \textbf{83.12} & \textbf{90.30} & \textbf{66.03} & \textbf{85.19} \\ \midrule \bottomrule
\end{tabular}
\end{table*}

%% file: sections/algorithms/algorithm2.tex
\begin{algorithm}[t!]
\caption{The Model Selection and Reuse Procedure.}
\label{alg:alg2}
\begin{algorithmic}[1]
    \STATE \textbf{Input:} Model hub \(\mathcal{M}\), model labels \(\{ S_m\}\), \(\mathrm{SDAG}\), reuse budget \(k\), target tasks \((\mathcal{X},\ \mathcal{Y})\)
    \STATE \textbf{Output:} Zero-shot prediction \(\{y'\}\)
    \STATE \textbf{Procedure:}
    \STATE {Semantic match \(\mathcal{Y}\) with \(\mathrm{SDAG}\) to obtain \(C_\mathcal{Y}\).\\ \(C_\mathcal{Y}=\{c\ |\ (y,c)\in \mathcal{Y}\times \mathrm{SDAG}, \forall y',c':\ \mathrm{sim}(y,c)\geq\mathrm{sim}(y',c)\wedge\mathrm{sim}(y,c)\geq\mathrm{sim}(y,c')\} \)}
    \STATE Employ CHCO Algorithm for each model \(F_m\).
    \FOR {\(F_m \in \mathcal{M}\)}
        \STATE \(\{\mathcal{L}^m_c\},\ \{\textbf{x}^m_c\} = \mathrm{CHCO}(S_m, C_\mathcal{Y})\)
        \STATE Construct \(|\mathcal{Y}|\) expert predictors \(F^m_c=\{(F_m,\textbf{x}^m_c)\}\)
    \ENDFOR
    \STATE Construct ensemble predictors with reuse budget \(k\).\\\(F_c^k=\{(F_m,\textbf{x}^m_c)|\mathcal{L}^m_c\leq \min(\delta, \text{top\(k\)-min}(\{\mathcal{L}_c^m\}_m)) \}\)
    \STATE Zero-shot predicting for \(\mathcal{X}\) with \(\{F_c^k\}\) by Eq.(\ref{eq.predict}).
    \RETURN zero-shot prediction \(\{y'\}\).
\end{algorithmic}
\end{algorithm}

%% file: sections/experiments.tex
\input{sections/figures/figure_exp2}

In this section, we conducte experiments on real datasets to validate the effectiveness of reusing expert models for zero-shot classification through MLL and the reuse efficiency of our method. Additionally, we showcase the scalability of the model hub and validate the superiority of the CHCO algorithm through an ablation study. We begin with a detailed description of the experimental setup.

\subsection{Experimental Setup}
\textbf{MLL Setting.} We constructe the MLL system as a reproducible benchmark as follows. The SDAG includes a total of 11,312 semantic nodes with a DAG structure extracted from WordNet. Each node collects up to 50 images from ImageNet. The text for each semantic node is formatted as ``Name: [synset name]\textbackslash{}nDescription: [synset description]", where the placeholders ``[]" are substituted with the corresponding synset's name and description. We use OpenAI text-embedding-3-large API to generate text embeddings for semantic matching. Forty-eight classification models are downloaded from the Huggingface community~\cite{wolf2020transformers} to form the basic model hub. These models vary in architecture, scale, output space, training data distribution, and classification purposes. We do not access their detailed functional descriptions or native information. All models are used without any fine-tuning or additional training.

\noindent\textbf{Datasets.} We conduct experiments on seven widely used image classification datasets across various domains, including Oxford-Pets~\cite{parkhi2012cats}, Food-101~\cite{bossard2014food101}, Caltech-101~\cite{fei-fei2004learning}, Oxford Flowers-102~\cite{nilsback2008automated}, and DOG, AWA and BRD from Meta Album~\cite{ullah2022metaalbum}.
The Oxford Pets includes 34 breeds of pet dogs and cats,
Food-101 features 101 types of food, 
Caltech-101 dataset contains 101 types of general objects, 
Oxford Flowers-102 encompasses 102 species of flowers.
Each sub-dataset of Meta Album contains 20 different species of dogs, large animals and birds respectively. 
We employed a basic zero-shot classification setup, utilizing the full test sets for evaluation and comparison.

\noindent\textbf{Compared Methods.} We compare our proposal with four candidates. CLIP~\cite{radford2021learning}, using the hand-crafted prompt ``a photo of a" served as a high-performance baseline. A random-based expert selection strategy is employed in the selection step, called Random\textsuperscript{MLL}. We also integrate state-of-the-art reusability assessment methods LogME~\cite{you2021logme} and LEEP~\cite{nguyen2020leep} into the MLL paradigm, denoted as LogME\textsuperscript{MLL} and LEEP\textsuperscript{MLL}, respectively. 
In the labelling step, we conduct balanced random sampling from the SDAG to generate a binary classification test for each class. LEEP and LogME assess each model and produce reuse metrics. We record metrics and classification heads with the minimum BCE loss as model labels, indicating the candidate expert predictors and their performance. In the selection step, we select predictors with the top reuse metrics for each class.

\noindent\textbf{Metrics.} We report the classification accuracy on the portion of the target task covered by the MLL system as Acc.(c) and the coverage rate as Cvr. Additionally, we provide the accuray of the entire task as Acc.(a) and the number of models reused in different reuse methods. 
Further implementation details can be found in the appendix.

\input{sections/tables/table2}

\subsection{The Comparison of Zero-Shot Performance }
We set the reuse budget \( k \) to 2, allowing up to two expert predictors to be assembled for each target class. The top section of Table \ref{tab:exp1} displays results of the accuracy Acc.(c) and the coverage Cvr., representing cases where the MLL system fully covers the target task. The bottom section of Table \ref{tab:exp1} shows the accuracy Acc.(a), illustrating the results of using MLL patch for entire tasks when the MLL system is not yet robust for the time being.
Results show that our method through the MLL proposal achieves high performance on the covered tasks, indicating that if the target task is fully covered by the MLL system, the selection and reuse of expert models can effectively address zero-shot tasks. 
Moreover, despite being in its preliminary stages and unable to address all tasks fully, the MLL system effectively reuses experts to enhance performance significantly compared to using a single general VLM, which highlights the usability of the MLL system, even if it is not yet developed sufficiently.

\subsection{The Comparison of Reuse Methods }
We vary the reuse budget \(k\) from 1 to 5 to compare our proposal against other reuse methods, LogME and LEEP.

\noindent\textbf{Efficacy.} Figure \ref{fig:exp2} shows that increasing reuse budgets generally boosts ensemble performance initially, but as more experts are incorporated, some less useful ones potentially lead to a decline in performance. Our proposal achieves robust and superior reuse results across different reuse budgets \(k\) by effectively selecting and reusing capable expert models for specific tasks. In contrast, candidate methods exhibit significant performance fluctuations with varying \(k\) and fail to outperform the CLIP baseline, indicating the inaccurate identification and ineffective reuse of experts.

\noindent\textbf{Efficiency.} The number of reused models reflects the efficiency of reuse methods. An efficient reuse scheme should utilize as few experts as possible while maintaining high performance. However, results indicate that LEEP\textsuperscript{MLL} and LogME\textsuperscript{MLL} often select more models during reuse, including some suboptimal or even useless ones. This leads to performance degradation and inefficiency. Involving unnecessary models wastes computational and time resources.

\subsection{The Scalability of the Model Hub}

We create a scenario where the model hub evolves from scratch to include all models. Figure \ref{fig:exp3} depicts the average performance across all tasks during this process. Results indicate that the reuse performance improves as the model hub scales. This suggests that a more diverse collection of expert models in the hub enhances its ability to handle a broader range of tasks. 
Our proposal with the patch starts with the CLIP baseline and steadily improves, achieving effective reuse of the expert models. When the hub is small and weak, we adaptively rely more on the general model to prevent performance degradation. As the hub grows, more experts are utilized to handle tasks better and enhance the zero-shot performance, demonstrating the feasibility of our proposal. In contrast, LEEP and LogME are more likely to reuse some ineffective experts, especially when the model hub is small. Worse, their underutilization of the model hub prevents them from surpassing the CLIP baseline

\input{sections/figures/figure_exp3}

\subsection{Ablation Stduy of CHCO}
We conducte an ablation study by replacing CHCO with a heuristic method (HEU) that selects one classification head with the minimum BCE loss as the expert predictor and corresponding reuse score for each class. To illustrate this more intuitively, we design a task called DvC, which merges the fine-grained classes in Oxford-Pets into two superclasses, dogs and cats. In a case study with MobileNet pre-trained on ImageNet-1k, downloaded from Huggingface, as shown in Table \ref{tab:exp4}, HEU selects two predictors for the two classes using only two classification heads and achieves 90.93\% accuracy on DvC. In contrast, CHCO selects two predictors by combining 61 useful heads through minimizing the discriminative loss, achieving 98.34\% accuracy, significantly outperforming HEU. We also report the comparison of reuse performance across other tasks.

Despite MobileNet not being specifically trained to categorize cats and dogs, effective zero-shot prediction on the DvC can still be achieved by reusing it through MLL.
In MLL, regardless of the original training purpose, as long as the selected expert predictors can discriminate target classes, the model has the reuse value for zero-shot prediction. This indicates that expert models can be used beyond their initial submission, allowing the MLL paradigm to flexibly address a broader range of tasks.

%% file: sections/figures/figure_exp2.tex
\begin{figure*}[t]
    \centering
    \includegraphics[width=\textwidth]{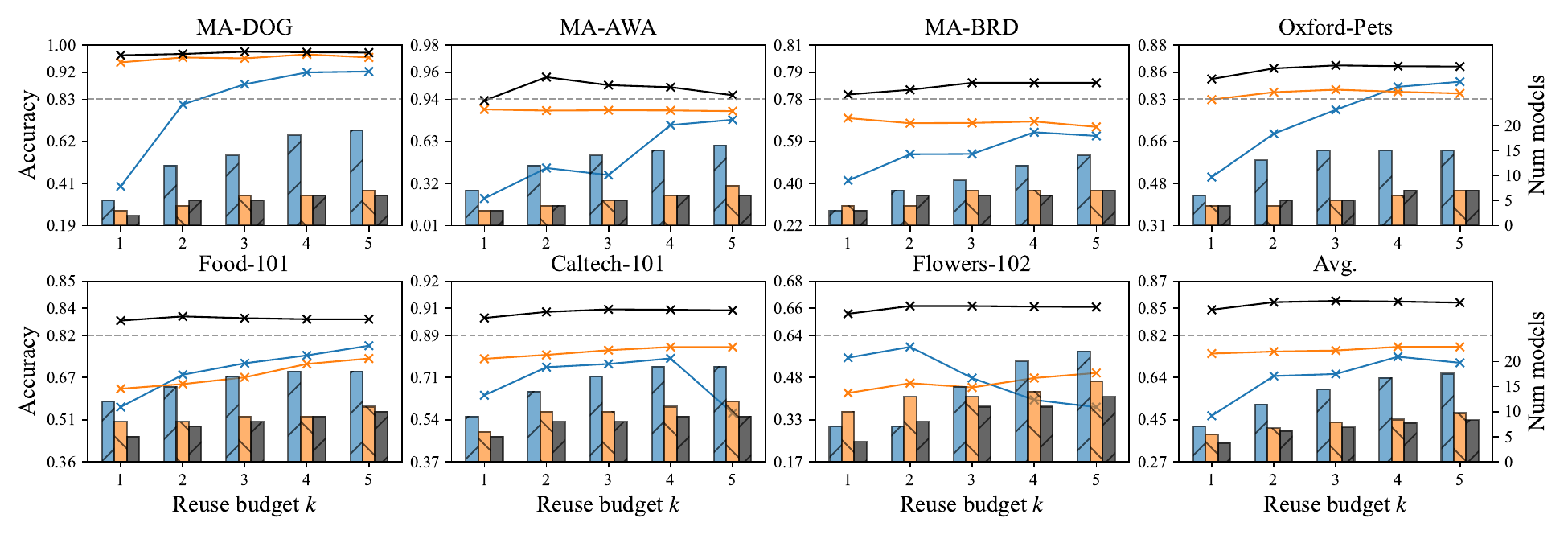} 
    \caption{The accuracy and the number of models used vary with the reuse budget \( k \). Ours, LogME\textsuperscript{MLL}, and LEEP\textsuperscript{MLL} are depicted in black, blue, and orange, respectively, with the CLIP baseline in the grey dashed line. Our proposal exhibits robust and superior performance, utilizing fewer expert models than candidate methods, which indicates higher reuse efficiency.}
    \label{fig:exp2}
\end{figure*}

%% file: sections/tables/table2.tex
\begin{table}[t!]
\centering
\small
\caption{Comparison of heuristics and CHCO. Expert predictors with multiple heads offer better reuse performance.}
\label{tab:exp4}
\begin{tabular}{@{}c|cccc@{}}
\toprule
 & \multicolumn{2}{c}{DVC-MobileNet} & \multicolumn{1}{c|}{Num Heads} & DVC \\ \midrule
HEU. & \multicolumn{2}{c}{90.93} & \multicolumn{1}{c|}{2} & 94.47 \\
CHCO & \multicolumn{2}{c}{98.34} & \multicolumn{1}{c|}{61} & 98.35 \\ \midrule \midrule
 & MA-DOG & MA-AWA & MA-BRD & Oxford-Pets \\ \midrule
HEU. & 96.58 & 89.63 & 73.38 & 85.20 \\
CHCO & \textbf{96.97} & \textbf{95.63} & \textbf{78.13} & \textbf{86.14} \\ \midrule
 & Food-101 & Caltech-101 & Flowers-102 & Avg. \\ \midrule
HEU. & \textbf{83.26} & 88.70 & 63.83 & 82.94 \\
CHCO & 83.12 & \textbf{90.12} & \textbf{66.03} & \textbf{85.16} \\ 
\bottomrule
\end{tabular}
\end{table}

%% file: sections/figures/figure_exp3.tex
\begin{figure}[t]
    \centering
    \includegraphics[width=\columnwidth]{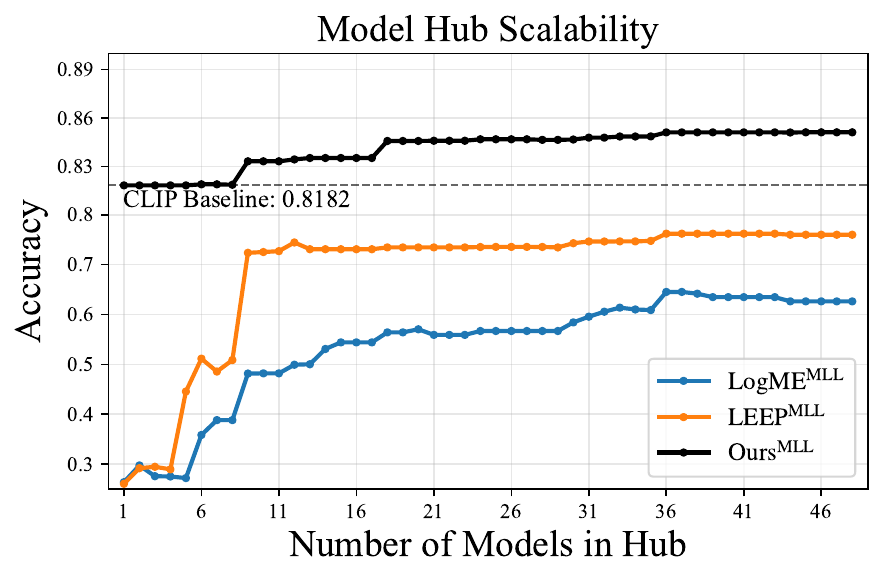} 
    \caption{The performance variation with the scaling of the model hub. Our method starts from the general baseline and steadily improves as the hub is continuously enhanced.}
    \label{fig:exp3}
\end{figure}